\documentclass[conference]{IEEEtran}
\IEEEoverridecommandlockouts
\usepackage{cite}
\usepackage{amsmath,amssymb,amsfonts}
\usepackage{graphicx}
\usepackage{textcomp}
\usepackage{xcolor}
\usepackage{url}
\usepackage{booktabs}
\usepackage{enumitem}
\usepackage{graphicx}
\usepackage{color}
\usepackage{multirow}
\usepackage{bbding}
\usepackage{tabularx}

\usepackage{graphicx}

\usepackage{algorithm}
\usepackage{algpseudocode}
\usepackage{amsmath}

\def\BibTeX{{\rm B\kern-.05em{\sc i\kern-.025em b}\kern-.08em
    T\kern-.1667em\lower.7ex\hbox{E}\kern-.125emX}}
\begin{document}

\newcommand{\modelname}{{MVMN}}

\title{Efficient Task Adaptation in Large Language Models via Selective Parameter Optimization \\
}

\author{
\IEEEauthorblockN{ Weijie Wan\textsuperscript{1}$^{*}$, Jiangjiang Zhao\textsuperscript{2}}
\IEEEauthorblockA{\textit{
\textit{\textsuperscript{1}Shortest Path Technology}\quad
\textsuperscript{2}China Mobile Communications Group Co.,Ltd}}
willjay@m.scnu.edu.cn
}

%\and
%\and
%\and
%\and
%\and

\maketitle

\begin{abstract}
Large Language Models (LLMs) have demonstrated excellent performance in general language understanding, generation and other tasks. However, when fine-tuning for specific domain tasks, the general knowledge accumulated in the pre-training phase is often partially overwritten or forgotten due to parameter updates, which severely limits the generalization ability and transferability of LLMs.
Traditional fine-tuning strategies mostly train on the entire parameter space, ignoring the heterogeneity of model parameters, that is, some parameters are extremely important for general tasks, while other parameters are more sensitive to specific tasks.
To alleviate the above problems, this paper innovatively proposes a parameter element importance evaluation method, which divides parameters into "core parameters" and "non-core parameters" by distinguishing the importance of parameters for general language ability tasks and specific domain tasks, and fixes the core parameters during fine-tuning, and only fine-tunes the non-core parameters.
Extensive experiments on scientific, medical and physical tasks using GPT-J and LLaMA-3 show that our method can mitigate catastrophic forgetting while enhancing the adaptability of the model.
\end{abstract}


\begingroup\renewcommand\thefootnote{*}
\footnotetext{Corresponding author.}
\endgroup

\section{Introduction}

\begin{figure}[t]
\centering
\includegraphics[width=0.45\textwidth]{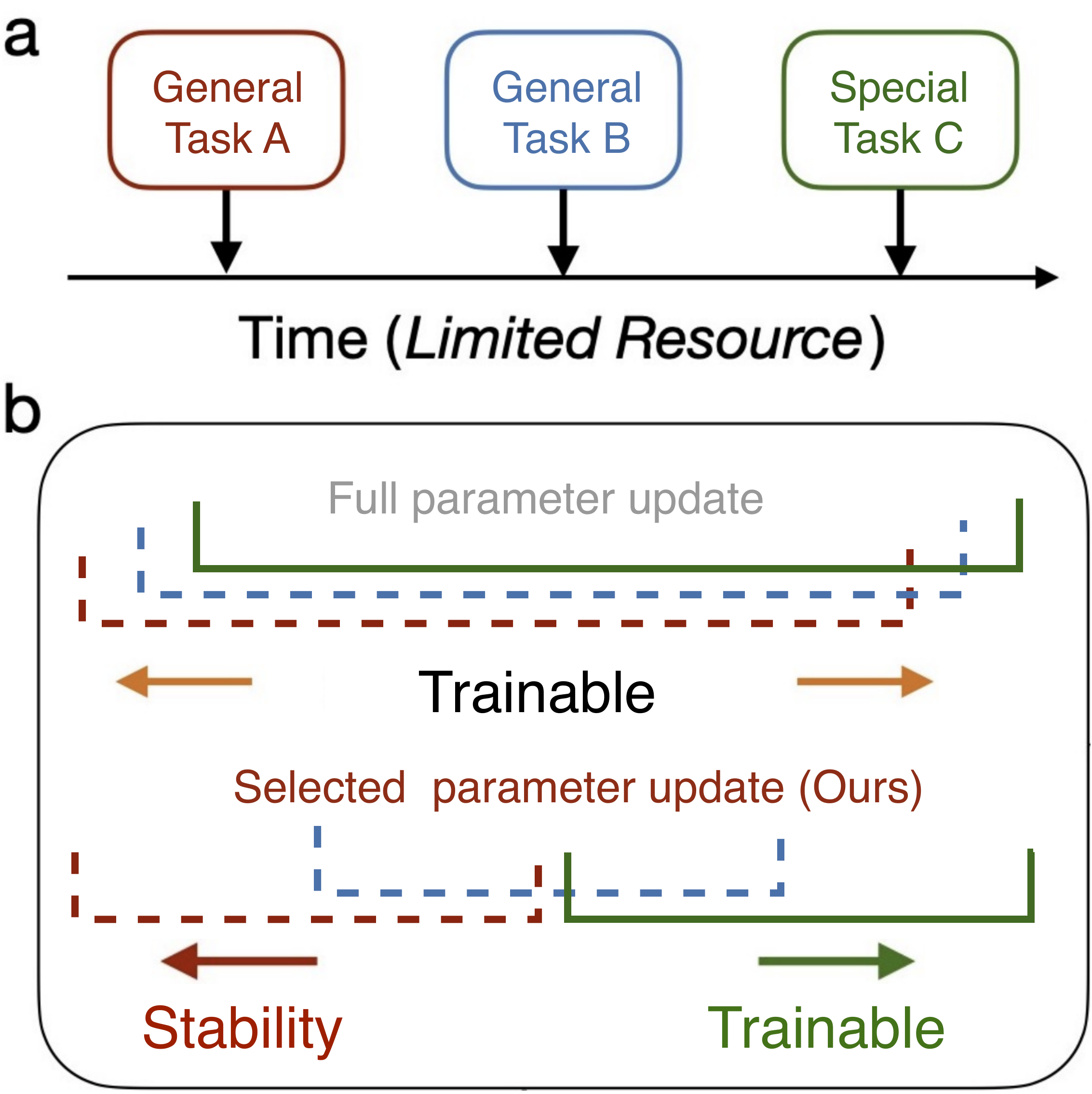}
\caption{Comparison of fine-tuning strategies for mitigating catastrophic forgetting. (a) illustrates the timeline of training data exposure in large language models (LLMs); (b) contrasts full-parameter fine-tuning with parameter-selective fine-tuning approaches.}
\label{fig1}
\end{figure}

Large language models (LLMs) have excellent general language capabilities due to pre-training on massive, multivariate datasets. This pre-training process gives LLMs powerful functions, making them play an important role in text generation, question-answering systems, and machine translation.
However, in actual application scenarios, in order to meet the needs of specific fields and tasks, such as professional literature analysis in the medical field and academic paper understanding in the scientific field, users often fine-tune LLMs. Although fine-tuning can significantly improve the performance of the model on specific tasks and make the model more suitable for applications in professional fields, it also brings a key problem that cannot be ignored - catastrophic forgetting.
From the figure \ref {fig1}, it can be clearly seen that full parameter fine-tuning will have catastrophic forgetting problems, that is, the model loses or covers the knowledge learned in the pre-training stage during fine-tuning. The root cause of this problem is that new field-specific training data usually cannot cover the rich and diverse, widely representative information in the pre-training corpus. During the fine-tuning phase, due to data limitations, the gradient will over-update the parameters that are critical to the general functions of LLM, which will cause the model's performance to drop sharply when dealing with tasks outside the fine-tuning domain. This problem seriously restricts the widespread application of LLM, greatly weakens its versatility and reusability between different fields, and makes it difficult for LLM to fully exert its original powerful advantages.

Therefore, to fully utilize the power of large language models (LLMs), solving the problem of catastrophic forgetting is the key. This requires a delicate balance: when LLMs learn new domain-specific expertise, they must also firmly retain the necessary general knowledge. This balance is particularly critical in the process of fine-tuning LLMs for specialized tasks. After all, in practical applications, both domain adaptability and generality are indispensable.
In dealing with the problem of catastrophic forgetting when fine-tuning LLMs, EWCLoRA\cite {ewclora} adopts a unique approach, which uses the Fisher matrix to evaluate the importance of parameters to the general capabilities of the model. However, this method has obvious drawbacks and is computationally expensive. Taking GPT-J-6B as an example, it takes up to 22 hours to calculate the Fisher matrix on an A800 device and also takes up 23GB of storage space. Moreover, as the scale of LLMs increases, these computing resource requirements will continue to rise, which undoubtedly limits its widespread application in practical scenarios.
Another method, RsLoRA\cite {rslora}, aims to stabilize the learning process by introducing a level-stable scaling factor. However, the actual effect is not satisfactory. It fails to effectively protect the general ability of the model as expected, performs poorly in maintaining the generality of the model, and fails to fully meet the needs of solving the catastrophic forgetting problem.

To address the above problems, in this paper we propose a two-stage training strategy based on "parameter importance" (PI). In the general capability stage, the model is first trained on a large-scale general training set, which can be regarded as a pre-training or large-scale fine-tuning process. During this period, the importance of each parameter to the general capability of the model is accurately calculated to prepare for the subsequent parameter differentiation. Then enter the task-specific fine-tuning stage, and distinguish between key parameters (Core Parameters) and non-core parameters (Non-core Parameters) based on the parameter importance calculated in the early stage. When fine-tuning for a specific task, only non-core parameters are updated. In this way, the model can retain the original generalization ability and quickly adapt to new fields, effectively avoiding the occurrence of catastrophic forgetting, and can improve the comprehensive performance of LLM in practical applications. And the method has been effectively verified after large-scale experiments on a variety of tasks such as science, physics, and medicine using mainstream large-scale language models (such as GPT-J and LLaMA-3). Experimental results show that this method effectively alleviates the catastrophic forgetting problem while improving the adaptability of large-scale language models to tasks in specific fields, and all performance indicators have achieved consistent improvements over the baseline.

The contributions of this paper are summarized as follows: First, in order to solve the problem that general knowledge is covered or lost during the fine-tuning of large language models (LLMs) in specific domains, an innovative parameter importance evaluation and controlled fine-tuning method is proposed, which effectively retains the pre-trained general knowledge and improves the adaptability of specific domain tasks.
Secondly, we propose a new parameter importance evaluation method, that is, a parameter importance calculation method is designed based on the general language ability task to quantify the contribution of model parameters to general tasks. By analyzing the dynamic influence of parameter gradients and updates, the core parameters in the model that are key to general capabilities are determined. In the process of fine-tuning specific domain tasks, the core parameters are fixed and only non-core parameters are adjusted, so as to achieve effective collaboration between specific domain task capabilities and general capabilities. 
Finally, to verify the effectiveness of this method, we conducted intensive experiments on multiple downstream tasks. The experimental results show that consistent improvements have been achieved in multiple specific domain tasks (such as medical text analysis, legal document processing, etc.), indicating that the model can retain both capabilities after select fine-tune the parameters.

\section{Related Work}

\subsection{Catastrophic Forgetting in Traditional Neural Networks}

Catastrophic forgetting refers to a model’s tendency to lose previously learned knowledge when trained on new tasks sequentially \cite{kemker2018measuring,liang2019adaptive,liang2019asynchronous,liu2023time}. This challenge is particularly pronounced in continual learning, where models are required to adapt without access to prior task data.
To address this issue, various strategies have been proposed. \textbf{Synaptic Intelligence (SI)} \cite{SI} dynamically estimates the importance of each parameter and penalizes updates to those crucial for past tasks, thereby preserving essential knowledge. Another widely adopted method is \textbf{Elastic Weight Consolidation (EWC)} \cite{Kirkpatrick2016OvercomingCF,xue2023dual,ma2022searching,wu2025unleashing,dai2025hope,gao2026decorl,li2026safety,lin2026parameterimportancestaticevolving}, which utilizes the Fisher Information Matrix to measure parameter sensitivity. It then adds a regularization term that restricts drastic updates to important parameters during new task training.
In parallel, \textbf{Learning without Forgetting (LwF)} \cite{li2017learning,zheng2022robust,song-etal-2022-improving-semantic,li2024local,li2024comateformer,liu2025stole,liu2026dpi,liu2025structural} addresses forgetting through knowledge distillation. By incorporating predictions from prior tasks during training, LwF maintains performance on older tasks while learning new ones. These methods have demonstrated generalizability across domains including vision and language. However, their effectiveness often diminishes on large-scale models due to scalability and memory constraints.

\subsection{Catastrophic Forgetting in LLMs and Parameter-Efficient Fine-Tuning (PEFT)}

The emergence of large language models (LLMs) \cite{touvron2023llama1,touvron2023llama2,zheng-etal-2022-robust,wang-etal-2022-dabert,qianadaptive,wang2026rethinking,wang2025not} has revolutionized NLP, but their massive parameter scale poses a significant challenge for domain adaptation. \textbf{Full-parameter fine-tuning} \cite{lv2023fullpeft,gui2018transferring,liu2024resolving} can lead to catastrophic forgetting due to indiscriminate updates across the network. Moreover, the computational and memory overhead makes it impractical in real-world deployments.
To alleviate this, researchers have proposed \textbf{Parameter-Efficient Fine-Tuning (PEFT)} strategies, such as \textbf{Low-Rank Adaptation (LoRA)} \cite{hu2021lora,wang2024lorapro,fei2022cqg,wu2025breaking,wu2026mmtablebench}, which insert trainable low-rank matrices while freezing the majority of model weights. While PEFT significantly reduces training cost and mitigates forgetting, performance degradation still occurs in sequential task settings.
Several hybrid approaches have been developed to further mitigate forgetting in PEFT. For example, \textbf{EWCLoRA} \cite{ewclora,chen2024s3prompt,xue2024question} integrates EWC into LoRA by estimating parameter importance via the Fisher Information Matrix. Although this improves knowledge retention, it introduces notable computational overhead, similar to challenges faced in generative modeling \cite{shen2024boosting}.
Another line of work, \textbf{Interpolation-based LoRA (I-LoRA)} \cite{ilora,liu2023local,wu2024tablebench,wu2025progressive,xue2026reason,xue2026supervised}, interpolates LoRA parameters during training to simulate a memory mechanism. While effective, it increases memory usage by maintaining dual LoRA modules throughout training, impacting deployment efficiency. Techniques that draw on modular or progressive fine-tuning also echo ideas from diffusion-based generation frameworks, which have shown success in selectively adapting task-specific components \cite{shen2023advancing,shen2024imagdressing}.
Recently, domain-specific applications of controllable generation \cite{shen2024imagpose}, story visualization \cite{shen2024boosting}, and long-term temporal modeling \cite{shen2025long} demonstrate that preserving structural consistency and prior knowledge across multiple conditions is both desirable and feasible. These insights could inspire future PEFT-based continual learning methods for LLMs.

\section{Background and Problem Definition}
\subsection{Background}
LLMs have demonstrated excellent language understanding, generation, and generalization capabilities by pre-training on massive and diverse general data. However, when trying to fine-tune the model for specific domain tasks (such as medical text analysis, legal document processing, etc.), it often faces the challenge of "catastrophic forgetting": the training data in the new domain often lacks the rich and diverse information covered by general pre-training, resulting in excessive updates to the key parameters of the model during the fine-tuning phase, thereby covering or losing the original general knowledge. To alleviate this problem, this paper proposes a two-stage fine-tuning strategy based on the importance of model parameter elements, which can improve the performance of specific domain tasks while retaining general capabilities.

\subsection{Problem definition}
\label{sec:problem_statement}
Let the set of trainable parameters of a large language model be $\mathbf{W} = \{w_1, w_2, \dots, w_N\},$
where $N$ is the number of parameters. To take into account both general capabilities and specific domain capabilities, we introduce two datasets:
\begin{itemize}
\item \textbf{General dataset} $\mathcal{D}_g$: contains large-scale and diverse corpora, used to consolidate or learn general language understanding capabilities;
\item \textbf{Domain dataset} $\mathcal{D}_s$: a sample set for the target domain or specific tasks, used to improve the level of specialization in the fine-tuning stage.
\end{itemize}
The traditional fine-tuning strategy will update all parameters $\mathbf{W}$ on $\mathcal{D}_s$, which may destroy the original general language ability. To resolve this conflict, we complete this through the following two modules:
\textbf{1) calculate and identify} the important parameters (i.e. key parameters) of the model on the general dataset $\mathcal{D}_g$;
\textbf{2) keep important parameters frozen} and only update other (non-core) parameters, so as to retain the general ability to the greatest extent while training the ability of specific fields.

\subsection{LoRA Tuning}
LoRA is a lightweight and parameter-efficient fine-tuning method that introduces low-rank decomposition into the weight matrix $\theta$ of a pretrained model. Only the newly added low-rank matrices $B$ and $ A $ are optimized, while the main weight $\theta_0$ remains frozen. The parameter at time $t$ during fine-tuning can be expressed as:
\begin{equation}
    \theta_t = \theta_0 + \delta_t, \quad \delta_t = B_tA_t
\end{equation}
where $ \theta_0 \in \mathbb{R}^{d \times d} $ are pretrained weights; $ B \in \mathbb{R}^{d \times r}, A \in \mathbb{R}^{r \times d}$ are the low-rank matrices with $ r \ll d $. The optimization objective of LoRA is given by:
\begin{equation}
    \mathcal{L}_{\text{LoRA}} = \mathcal{L}(y, f(x; \theta(t)))
\end{equation}
where $ \mathcal{L} $ is the task-specific loss function. Although LoRA achieves parameter efficiency and training effectiveness, it suffers from catastrophic forgetting, where fine-tuning specific tasks hurts the general ability.

\section{Method}
\subsection{Parameter Importance Calculation}
\label{sec:param_importance}

\subsubsection{General ability training}
On a general dataset $\mathcal{D}_g$, the training objective of a large language model can usually be defined as minimizing the expected value of a certain loss function, such as the cross entropy loss of language model training.
Let $(x,\, y)$ represent a sample in a general dataset and its corresponding label (or target sequence), then the loss of general training can be abstractly expressed as
\begin{equation}
\label{eq:general_loss}
\mathcal{L}_g (\mathbf{W}) \;=\; \mathbb{E}_{(x,y)\,\sim\,\mathcal{D}_g}\bigl[\ell\bigl(F_{\mathbf{W}}(x),\,y\bigr)\bigr],
\end{equation}
Where $F_{\mathbf{W}}(\cdot)$ represents the model forward calculation function with $\mathbf{W}$ as a parameter, and $\ell(\cdot,\cdot)$ represents the specific loss metric (such as cross entropy). By performing several rounds of training or fine-tuning on $\mathcal{D}_g$, the initial model weights $\mathbf{W}_{\text{general}}$ with general semantic understanding (or generation) capabilities can be obtained.

\subsubsection{Parameter importance measurement}
%After completing the general capability training, it is necessary to measure the importance of each parameter to the general capability of the model so as to identify and protect it in the fine-tuning stage. For parameter $w_i$, common evaluation methods include:
After completing the general ability training phase, measuring the importance of each parameter to the general ability of the model becomes a crucial step. As mentioned above, when fine-tuning a large language model (LLM), the catastrophic forgetting problem seriously affects the performance of the model in tasks in different fields, and accurately measuring the importance of parameters is a key step to effectively solve this problem. The core purpose of parameter importance measurement is to clearly identify which parameters play a key supporting role in the general ability of the model, so that in the subsequent fine-tuning phase, these key parameters can be protected in a targeted manner to prevent them from being improperly updated during the fine-tuning process, causing the model to lose important pre-training knowledge. In this section, we try to evaluate the importance of parameters based on the gradient evaluation method. By calculating the gradient changes of parameters during the training process, we can intuitively reflect the sensitivity of the parameters to the model loss function. If the gradient change of a parameter is large, it means that it has a significant impact on the overall performance during the learning process of the model, and may have an important contribution to the general ability of the model; conversely, parameters with small gradient changes are relatively less important. The details are as follows:
\textbf{Gradient-based Importance}
Let $\ell\bigl(F_{\mathbf{W}}(x),\,y\bigr)$ be the loss function of a single sample, then the importance of $w_i$ can be defined as the expectation of the absolute value of the loss gradient:
\begin{equation}
\label{eq:grad_importance}
I_i \;=\; \mathbb{E}_{(x,y)\,\sim\,\mathcal{D}_g}\Bigl[\;\bigl|\
abla_{w_i}\,\ell\bigl(F_{\mathbf{W}}(x),\,y\bigr)\bigr|\Bigr].
\end{equation}
After measuring the importance of parameters through the above gradient-based evaluation method, we will get a value to represent its importance. The larger the value, the more significant the impact on the loss function will be even if there is a very small change, which means that it plays a more critical role in the general ability of the model.

After such a measurement strategy, we can get a set of importance scores $I_1,I_2,\dots,I_N$. Based on this set of scores, in order to more accurately screen out key parameters, we further set a threshold $\theta$. Parameters that meet the threshold are judged as key parameters that have an important impact on the general ability and performance of the model. These parameters will be protected in the subsequent fine-tuning process to avoid catastrophic forgetting,such as
\[
I_i \;\ge\;\theta
\]
as key parameters, and the rest as "non-core parameters". Let the former set be $\mathbf{W}^K$, and the latter set be $\mathbf{W}^N$, then:
\[
\mathbf{W}^K \;=\;\{ w_i\mid I_i \ge \theta\},\quad
\mathbf{W}^N \;=\;\mathbf{W} \setminus \mathbf{W}^K.
\]

\subsection{Specific domain fine-tuning}
\label{sec:domain_finetune}

When fine-tuning a domain-specific task, in order to adapt to the task requirements of the specific domain, the model needs to update some of its parameters to capture domain-related features and semantics. However, since domain tasks and general pre-training tasks may have large differences in data distribution and objectives, this direct optimization usually leads to over-adjustment of general ability parameters, which in turn leads to **knowledge forgetting problem**, that is, the general language ability accumulated in the pre-training stage is overwritten or lost.

In this section, we designed a controlled fine-tuning method based on the division results of \textbf{parameter importance}. During the fine-tuning process of specific domain tasks, only the non-core parameter set $\mathbf{W}^N$ is updated, while the core parameter set $\mathbf{W}^K$ is fixed, so as to take into account both domain adaptation and general ability retention.

Specifically, for downstream tasks in specific domains, we first define the loss function for optimization. Assume that the dataset for domain fine-tuning is $\mathcal{D}_s$, then the task objective is to minimize the domain loss $\mathcal{L}_s(\mathbf{W})$ on the dataset. The general form of the loss function can be expressed as:
\begin{equation}
\label{eq:domain_loss}
\mathcal{L}_s (\mathbf{W}) \;=\;\mathbb{E}_{(x',y')\,\sim\,\mathcal{D}_s}\Bigl[\ell\bigl(F_{\mathbf{W}}(x'),\, y'\bigr)\Bigr],
\end{equation}
In the original LLM training and fine-tuning process, the commonly used practice is to perform gradient descent updates on all parameters in the model parameter matrix $\mathbf{W}$. This update method allows the model to gradually learn the characteristics and rules in the data when dealing with routine tasks, thereby achieving a certain degree of performance improvement. However, when the model faces a specific task that is significantly different from the general data distribution, since all parameters are updated by gradient descent indiscriminately, those parameters that are critical to the general ability of the model are likely to be disturbed by the specific task data, which in turn causes the model to conflict with the original general ability in the process of learning specific task knowledge. For example, when fine-tuning the model in the medical field, parameter updates related to new medical terms and expertise may destroy the general semantic understanding ability accumulated by the model in daily language understanding tasks, causing the model to perform significantly worse when dealing with general text in non-medical fields.
In order to effectively retain the general ability of the model so that it can adapt to specific tasks without losing the key knowledge learned in the pre-training stage, we use the following solution:
\begin{equation}
\begin{aligned}
w_i \,\leftarrow\,
\begin{cases}
w_i, & \text{if } w_i \in \mathbf{W}^K,\\
w_i \;-\;\eta \,\nabla_{w_i}\,\mathcal{L}_s(\mathbf{W}), & \text{if } w_i \in \mathbf{W}^N.
\end{cases} \\
\forall\,i=1,\dots,N,
\end{aligned}
\label{eq:update_selective}
\end{equation}
where $\eta$ is the learning rate. The core parameters $\mathbf{W}^K$ are kept frozen (not updated) during the fine-tuning phase to avoid compromising general capabilities; the non-core parameters $\mathbf{W}^N$ are fully involved in training to absorb domain-specific knowledge.

\begin{figure}[t]  % 使用 t 表示靠页面顶部
\renewcommand{\arraystretch}{1.2}
\vspace{5mm}
\hrule
\noindent
\textbf{Algorithm 1:} Two-stage fine-tuning strategy via parameter importance estimation and selective freezing
\vspace{1mm}
\hrule
\begin{itemize}
    \item \textbf{Inputs:} \\
    \quad Initial LLM parameters $\mathbf{W}$; \\
    \quad General-domain dataset $\mathcal{D}_g$; \\
    \quad Specific-domain dataset $\mathcal{D}_s$; \\
    \quad Importance threshold $\theta$, learning rate $\eta$, number of iterations, etc.
    
    \item \textbf{Output:} \\
    \quad Fine-tuned model parameters $\mathbf{W}'$
\end{itemize}

\vspace{1mm}
\begin{enumerate}[leftmargin=1.5em]
    \item \textsf{General-domain training:} Optimize the loss in Eq.~\eqref{eq:general_loss} on $\mathcal{D}_g$ to obtain the general model $\mathbf{W}_{\text{general}}$.
    
    \item \textsf{Estimate parameter importance:} For each parameter $w_i \in \mathbf{W}_{\text{general}}$, compute its importance score $I_i$ using Eq.~\eqref{eq:grad_importance}.
    
    \item \textsf{Separate key and non-key parameters:}
    \[
    \mathbf{W}^K = \{w_i \mid I_i \ge \theta\}, \quad
    \mathbf{W}^N = \mathbf{W} \setminus \mathbf{W}^K
    \]
    
    \item \textsf{Domain-specific adaptation:} Fine-tune only the non-key parameters on $\mathcal{D}_s$ by minimizing the domain loss in Eq.~\eqref{eq:domain_loss}:
    \[
    w_i \leftarrow
    \begin{cases}
        w_i, & \text{if } w_i \in \mathbf{W}^K \quad \text{(frozen)} \\
        w_i - \eta \nabla_{w_i} \mathcal{L}_s(\mathbf{W}), & \text{if } w_i \in \mathbf{W}^N
    \end{cases}
    \]
    
    \item \textsf{Return fine-tuned weights:} $\mathbf{W}' = \{w_i \mid w_i \in \mathbf{W}\}$.
\end{enumerate}
\vspace{1mm}
\hrule
\vspace{2mm}
\end{figure}

\subsection{Method Summary}
The overal method can be summarized as a two-stage process, as shown in the algorithm.

And the fine-tuning method based on "parameter importance" calculation proposed has the following advantages:
 \textbf{Retain general knowledge:} By analyzing and fixing the parameters that play a key role in comprehensive capabilities on a general dataset, the damage to general capabilities during fine-tuning is greatly reduced;
\textbf{Flexible adaptation to new tasks:} Non-core parameters still have sufficient update space, which can enable the model to achieve effective bias and adaptation in specific fields;
 \textbf{Wide range of application:} Importance measurement can be based on different measurement criteria (absolute value of gradient, incremental loss, etc.), which is feasible for various large models and multiple tasks;
 \textbf{Easy to implement:} Only one additional statistics and index of parameter importance is required on the existing fine-tuning process, and there is no need to change the original network structure or add additional adaptation layers.

\begin{table*}[t]
  \centering
  \renewcommand\arraystretch{1.5}
  \caption{Evaluation of general and domain capabilities on SciQ, PiQA, and MedMCQA. 
  (PPL$\downarrow$: lower is better; Acc$\uparrow$: higher is better)}
  \resizebox{\linewidth}{!}{
    \begin{tabular}{lcccccc|cccccc}
    \toprule
    \multirow{2}{*}{} & \multicolumn{6}{c|}{\textbf{LLaMA-3}} & \multicolumn{6}{c}{\textbf{GPT-J}} \\
    \cmidrule(lr){2-7} \cmidrule(lr){8-13}
     & \multicolumn{2}{c}{SciQ} & \multicolumn{2}{c}{PiQA} & \multicolumn{2}{c|}{MedMCQA} & \multicolumn{2}{c}{SciQ} & \multicolumn{2}{c}{PiQA} & \multicolumn{2}{c}{MedMCQA} \\
     & PPL$\downarrow$ & Acc$\uparrow$ & PPL$\downarrow$ & Acc$\uparrow$ & PPL$\downarrow$ & Acc$\uparrow$ & PPL$\downarrow$ & Acc$\uparrow$ & PPL$\downarrow$ & Acc$\uparrow$ & PPL$\downarrow$ & Acc$\uparrow$ \\
    \midrule
    Base & 4.94 & 95.10 & 4.94 & 48.53 & 4.94 & 18.50 & 3.28 & 91.60 & 3.28 & 49.13 & 3.28 & 21.30 \\
    LoRA($\mu$) & 5.05 & 96.20 & 5.43 & \underline{48.75} & 5.04 & 53.69 & \underline{3.43} & \underline{96.50} & 3.54 & 50.16 & 3.49 & \textbf{38.35} \\
    LoRA($\nu+\mu$) & 5.31 & 96.10 & 5.58 & 46.91 & 5.15 & 53.12 & 3.48 & 96.50 & \underline{3.45} & \underline{50.44} & 3.41 & \underline{37.13} \\
    RSLoRA & 5.28 & \underline{96.50} & 5.71 & 47.50 & 5.24 & 51.92 & 3.50 & 96.20 & 3.65 & 49.62 & \textbf{3.31} & 35.69 \\
    EWCLoRA & \underline{4.88} & 96.30 & \underline{4.98} & 48.45 & \underline{4.79} & \textbf{56.39} & 3.38 & 96.10 & 3.47 & 49.40 & 3.38 & 36.48 \\
    \textbf{Ours} & \textbf{4.74} & \textbf{97.08} & \textbf{4.92} & \textbf{51.16} & \textbf{4.62} & \underline{55.88} & \textbf{3.34} & \textbf{96.92} & \textbf{3.38} & \textbf{50.49} & \underline{3.36} & 36.18 \\
    \bottomrule
    \end{tabular}
  }
\label{tab:mainresult}
\end{table*}

\section{Experiments}
\subsection{Backbone LLM}
We evaluate the proposed method using two mainstream LLM:
(1) \textit{GPT-J} \cite{gpt-j} is a 6-billion-parameter transformer-based model developed by EleutherAI. It is pretrained on a diverse corpus of texts, making it suitable for a wide range of natural language understanding and generation tasks.
(2) \textit{LLaMA-3} \cite{dubey2024llama3} is the second-generation open-source language model developed by Meta AI. It is designed with enhanced efficiency and scalability, offering state-of-the-art performance across various benchmarks. 
% \item MiniCPM3 is a lightweight Chinese language model designed for efficiency and adaptability. It leverages pretraining on extensive multilingual corpora and is tailored for tasks requiring domain adaptation. 
These models vary in architecture and parameter count, enabling a robust evaluation of our method.

\subsection{Tasks, Metrics and Hyperparameters}
\textbf{$\nu$ Task (General Ability):}
The $\nu$ task focuses on learning which parameters are important for general tasks. 
Following previous work \cite{ewclora}, we take Pile \cite{gao2020pile} as the evaluation datasets for LLM general ability. LoRA is applied to fine-tune the model on the $\nu$ task, and parameter importance for Synaptic Intelligence (SI) is recorded during this stage.

\textbf{$\mu$ Task (Domain Ability):}
The $\mu$ task evaluates the ability to adapt to specific tasks while mitigating catastrophic forgetting of general knowledge. We select three representative tasks:
(1) \textit{Medical task:} Using the MedMCQA dataset \cite{pal2022medmcqa} related to medical information extraction and understanding. 
(2) \textit{Scientific task:} Using the SciQ dataset \cite{welbl2017sciq} related to scientific  reasoning.
(3) \textit{Physics task:} Using the PiQA dataset \cite{bisk2020piqa} related to physical reasoning and problem-solving.

The LLMs selected for our experiments are GPT-J-6B and Llama 3.2-3B. The batch size is set to 20, and the learning rate is set to 8e-4. The rank for LoRA fine-tuning is set to 8, with the LoRA alpha value set to 32. Both the $\nu$ and $\mu$ tasks are trained for 5 epochs.

\subsection{Baseline Methods}

To comprehensively evaluate the effectiveness of our proposed method, we compare it against several state-of-the-art and widely adopted approaches. These baselines provide a strong foundation for assessing the trade-offs between domain-specific adaptation and general language ability preservation:
(1) \textit{Base}: This serves as the simplest baseline, where the pre-trained model is directly used without any fine-tuning. While this approach ensures no catastrophic forgetting, its performance on domain-specific tasks is inherently limited due to the lack of task-specific adaptation.
(2) \textit{LoRA($\mu$)} \cite{hu2021lora}: In this approach, the model is fine-tuned using only data from the $\mu$ task (domain-specific task). LoRA employs low-rank matrices to adapt the model parameters efficiently, reducing computational overhead while achieving competitive performance on domain-specific tasks. 
(3) \textit{LoRA($\nu + \mu$)}: This sequential fine-tuning strategy first adapts the model using data from the $\nu$ task (general task), followed by additional fine-tuning on the $\mu$ task (domain-specific task). By incorporating general task data before domain-specific adaptation, this method aims to mitigate catastrophic forgetting. 
(4) \textit{EWCLoRA} \cite{ewclora}: This method integrates Elastic Weight Consolidation (EWC) with LoRA to address catastrophic forgetting. Specifically, EWCLoRA computes the Fisher Information Matrix to estimate parameter importance and applies regularization constraints during fine-tuning. This ensures that important parameters for previously learned tasks are preserved while adapting to new tasks. 
(5) \textit{RSLoRA}: An enhanced variant of LoRA, RSLoRA modifies the scaling factor of the low-rank matrices to prevent gradient collapse during training. This modification enables better fine-tuning performance with higher-rank adapters while maintaining the same inference cost.

\section{Results and Analysis}
\subsection{Accuracy Comparision on General and Domain}
As shown in Table \ref{tab:mainresult} , our proposed method can significantly better preserve general language ability while achieving domain-specific task accuracy comparable to or even better than existing methods, as reflected in the lowest perplexity level (PPL) indicator. From the results, it can be seen that our method has achieved a good balance between domain adaptation tasks and general ability preservation. Unlike previous methods that only focus on domain performance optimization, our research provides a more comprehensive solution and establishes a new standard for model design that takes into account both domain task performance and general language ability.

% \subsection{Loss Curve Analysis}
As shown in Figure, we illustrate the loss curves for GPT-J and LLaMA-3 during their training process on three datasets. The total loss consists of two components: the task loss $\mathcal{L}_{\text{task}}$ and the regularization loss $\mathcal{L}_{\text{reg}}$, which together determine the model's optimization objectives in a weighted manner. From the curves, it can be observed that the task loss exhibits a consistent downward trend, reflecting the model's growing ability to adapt to the specific task. In contrast, the regularization loss $\mathcal{L}_{\text{reg}}$ displays a dynamic pattern, initially increasing and then decreasing as the training progresses.
According to the definition in Equation \ref{eq:update_selective}, $\mathcal{L}_{\text{reg}}$ measures the deviation between the model parameters $\theta_{\nu}$ learned after training on a prior task $\nu$ and the parameters $\theta_{\mu}$ learned during the current task $\mu$. In the early stages of training for task $\mu$, since the model parameters have not yet undergone substantial updates, the parameters of task $\mu$ remain relatively close to those of task $\nu$, resulting in a zero regularization loss. However, as the task loss $\mathcal{L}_{\text{task}}$ drives parameter updates, the deviation between $\theta_{\nu}$ and $\theta_{\mu}$ gradually increases, causing the regularization loss to rise. As training progresses further, the model converges to a more optimal set of parameters that balances the minimization of both the total loss and the task loss, leading to a coordinated reduction in both components. 
This mechanism ensures that the model optimizes task-specific performance while simultaneously retaining knowledge from prior tasks to the greatest extent possible. The regularization term acts as a constraint during training, introducing a penalty based on parameter deviation to guide the model in making moderate adjustments while learning new tasks. This effectively mitigates conflicts and catastrophic forgetting between tasks. The dynamic constraint mechanism is designed to strike a balance between task adaptability and knowledge retention, thereby enhancing the overall stability and effectiveness of the model.

\subsection{Complexity Comparision}

\textbf{Time Complexity:}
To quantitatively evaluate the time complexity, we compared the importance computation overhead of ALoRA and EWCLoRA on an A800 GPU. The time complexity of EWCLoRA is based on the Fisher matrix computation process described in its original paper. Specifically, the experiment used 20,000 randomly sampled samples from the Pile dataset and set the maximum batch size to 8, strictly following the standardized experimental configuration. In contrast, the time complexity of ALoRA is based on a customized optimization process, which only requires 5 training cycles to calculate parameter importance. This number of cycles was determined empirically through comprehensive experimental trade-offs between training stability and performance.
Experimental results show that for the two models GPT-J-6B and LLaMA-3-3B, EWCLoRA requires \textbf{27.17 hours} and \textbf{25.97 hours} to complete the Fisher matrix calculation. Our ALoRA method only takes \textbf{1.15 hours} and \textbf{1.19 hours} to complete the same task. The experiment further confirms the time efficiency of ALoRA in importance calculation, greatly reducing the overhead of relying on large datasets for complex calculations, and can adapt to a variety of training scenarios more quickly.

\textbf{Storage Memory:}
For the evaluation of storage memory consumption, we compared and analyzed the memory overhead required by the two methods when calculating and storing the parameter importance matrix. EWCLoRA relies on the Fisher matrix in parameter importance calculation, which needs to be calculated and stored in advance through the Pile dataset. According to the description of the original EWCLoRA paper, the Fisher matrix of the GPT-J-6B model occupies about \textbf{22.65 GB} of memory, while for the LLaMA-3-3B model, this memory requirement is \textbf{11.97 GB}. These results are based on the calculation process and formula of the Fisher matrix in the original method.
In contrast, the ALoRA architecture significantly reduces the storage memory requirements. For example, ALoRA only requires \textbf{3.5 GB} and \textbf{1.3 GB} of storage memory for the GPT-J-6B and LLaMA-3-3B models, respectively. This memory advantage is mainly due to ALoRA's lightweight design, which avoids the storage of high-dimensional Fisher matrices during the calculation of the importance matrix, thus providing a significantly more resource-efficient solution.
    % EWC: GPTJ 22.65G; 11.96G
    % Ours:GPTJ 3.5G; LLaMA 1.3GF

A comprehensive comparison of time and storage memory shows that although the computational process of EWCLoRA can capture the importance of parameters, its efficiency and resource utilization in large-scale data processing face significant bottlenecks. In contrast, ALoRA overcomes these shortcomings and provides a more efficient computing strategy, which not only significantly reduces time and storage costs, but also shows stronger applicability in environments with limited hardware and data scale. This shows that our ALoRA method is a more practical and efficient solution, opening up new possibilities for model training optimization in the context of low resource consumption.

\section{Conclusion}
In this paper, we propose an innovative method to effectively reduce the catastrophic forgetting of general capabilities when fine-tuning large language models in specific domains. The core parameters are fixed by evaluating the importance of parameters based on general capability tasks, and the damage to general capabilities is reduced by focusing only on non-core parameter training during specific task training. In order to evaluate the effectiveness of the proposed method, we conducted intensive experiments on multiple public evaluation sets. Experimental results show that this method can not only achieve significant improvements in multiple specific domain tasks, but also retain the general capabilities of the model to the greatest extent.

\end{document}